\documentclass[runningheads]{llncs}
\usepackage{graphicx}
\usepackage{amsfonts}
\usepackage{bbm}
\usepackage[colorlinks=true,linkcolor=blue, citecolor=blue, urlcolor=blue]{hyperref}
\usepackage{hhline}
\usepackage{ulem}
\usepackage{enumitem}
\usepackage{amsmath}
\usepackage[table]{xcolor}
\usepackage{booktabs}
\usepackage{makecell}
\usepackage{multirow}
\usepackage{xcolor}
\usepackage{siunitx}   
\renewcommand{\labelitemi}{$\circ$}
\definecolor{sig}{RGB}{255,255,210}
\definecolor{intracol}{RGB}{248,245,255}
\usepackage{pifont}
\usepackage{marvosym}
\newcommand{\Cmark}[1][green!60!black]{\textcolor{#1}{\ding{51}}} 
\newcommand{\Xmark}[1][red!70!black]{\textcolor{#1}{\ding{55}}}

\begin{document}

\title{IntraStyler: Intra-Domain Style Synthesis for Cross-Modality MRI Domain Adaptation}



\author{Han Liu\inst{1,2}\Letter\and
Yubo Fan\inst{2} \and
Hao Li\inst{2} \and
Dewei Hu\inst{3} \and
Daniel Moyer\inst{2} \and
Zhoubing Xu\inst{4} \and
Benoit Dawant\inst{2} \and
Ipek Oguz\inst{2}}

\institute{
Siemens Healthineers, Princeton, NJ, USA\and
Vanderbilt University, Nashville, TN, USA\and
Mayo Clinic, Rochester, MN, USA\and
Johnson \& Johnson Innovative Medicine, Titusville, NJ, USA
\\
\email{han.liu@siemens-healthineers.com}}
\authorrunning{H. Liu et al.}
\titlerunning{IntraStyler}

\maketitle              
\begin{abstract}

Segmentation of vestibular schwannoma and cochlea from T2 MRI is 
clinically important yet annotation-intensive. Domain adaptation (DA) has been widely adopted to bridge the gap 
between labeled contrast-enhanced T1 and unlabeled T2 datasets. 
While existing methods focus on cross-domain alignment, intra-domain 
variability within the target domain remains largely overlooked. 
Images from the same domain may vary substantially due to different 
scanners, field strengths, and acquisition protocols. Ignoring 
this variability produces homogeneous synthetic images that limit 
the generalizability of downstream segmentation models. To address this, we propose \textbf{\textit{IntraStyler}}, a 3D 
unpaired image translation method that automatically discovers 
fine-grained intra-domain styles without any predefined sub-domains, and synthesizes diverse target domain images using per-image 
style references. To this end, we design a 3D style encoder trained 
with a novel contrastive learning objective to extract 
style-only embeddings disentangled from anatomy. IntraStyler is built upon 
the 1st place CrossMoDA challenge solution and further advances it, generating more diverse synthetic data and 
achieving more reliable downstream segmentation. Code is available 
at \url{https://github.com/MedICL-VU/IntraStyler}.

\end{abstract}

\section{Introduction}

Vestibular schwannoma (VS) is a benign tumor arising from the 
vestibular nerve, and accurate segmentation of VS and cochlea from 
MRI is essential for treatment planning and follow-up. While 
contrast-enhanced T1 (ceT1) MRI is the standard modality, 
high-resolution T2 imaging offers a lower-risk and more 
accessible alternative~\cite{coelho2018mri}. However, 
annotating T2 images is costly and time-consuming, motivating 
domain adaptation (DA) approaches that transfer knowledge from 
labeled ceT1 to unlabeled T2 without target domain annotations~\cite{dorent2023crossmoda,challenge}. 
One of the most effective DA strategies is image-level domain 
alignment~\cite{dong2021unsupervised,liu2023learning,shin2022cosmos}, where source domain images are synthesized into the 
target domain via unpaired image 
translation~\cite{zhu2017unpaired,park2020contrastive}, and the 
synthetic images are used with source domain labels to train the 
segmentation model. Classical DA assumes images within the same domain share the same 
distribution. However, in many real-world medical datasets, images 
from the same domain (e.g., T2 MRI) may span multiple sub-domains 
arising from differences in scanners or acquisition 
protocols, also known as \textit{\textbf{intra-domain}} 
variability. If DA methods fail to capture this variability and 
produce only homogeneous synthetic images, the downstream 
segmentation model will have poor generalizability across the 
target domain.

\begin{figure}[t]
\includegraphics[width=1\columnwidth]{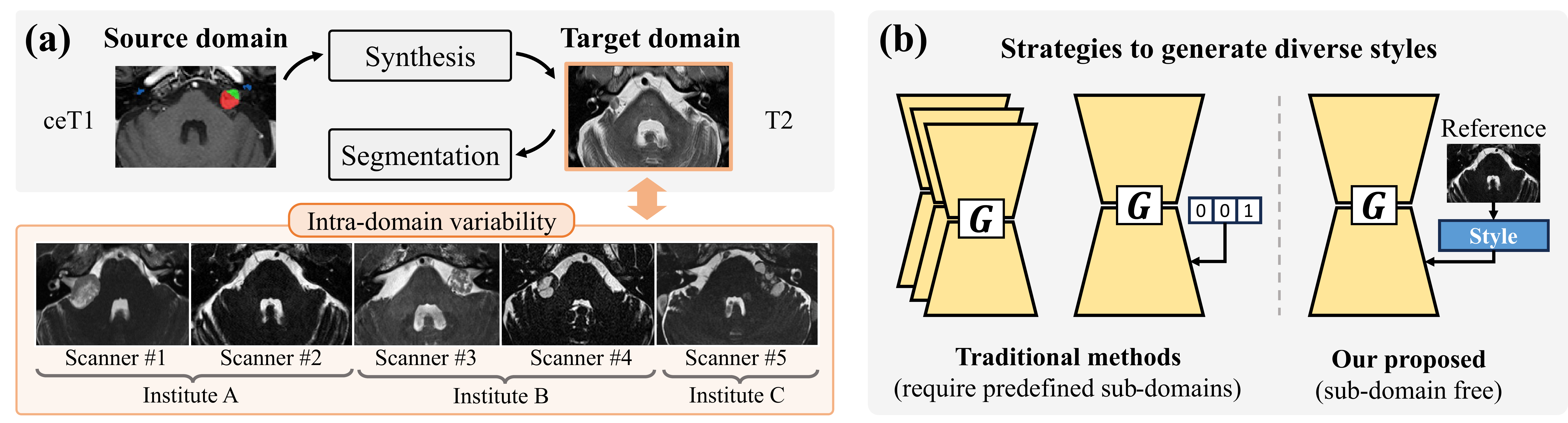}
\centering
\caption{\textbf{(a)} Intra-domain variability: images from the same domain 
exhibit diverse appearances across institutes and scanners. \textbf{(b)} Comparison of style generation strategies.}
\label{fig1}
\end{figure}

Most existing methods require \textbf{predefined sub-domains} to synthesize 
diverse styles. This assumption has two limitations in practice. \underline{First}, 
sub-domain labels are not always available. \underline{Second}, coarse sub-domain definitions 
fail to capture fine-grained appearance variation. As shown in Fig.~\ref{fig1}, previous studies use \textit{institutes} 
as sub-domains, yet images from the same institute may still differ substantially 
due to varying \textit{scanners} or \textit{protocols}.~\cite{liu2022enhancing,zhuang20233d} train 
separate synthesis networks per source and target sub-domain pair, which is highly 
inefficient and degrades when sub-domain data is scarce. 
~\cite{liu2023learning,han2023fine,choi2020stargan,li2024deep} train a unified network conditioned on predefined sub-domains, which still fundamentally depends on their availability and accuracy. It is therefore desirable to develop a method that captures intra-domain variability without any predefined sub-domains.


In this paper, we present \textbf{\textit{IntraStyler}}, an unpaired 
image translation method that captures diverse intra-domain styles 
without requiring any sub-domain labels. Our key insight is that each 
target domain image implicitly encodes its own fine-grained style and 
can therefore serve as a style reference at inference time. To extract 
style information disentangled from anatomy, we design a 3D 
style encoder trained with a novel contrastive learning objective. 
Built upon the 1st place CrossMoDA solution~\cite{liu2023learning}, 
IntraStyler demonstrates that (1) sub-domain-free style discovery yields 
more diverse synthetic images and (2) improves downstream segmentation 
generalizability. Our contributions are as follows:

\renewcommand{\labelitemi}{$\bullet$}
\begin{itemize}
    \item We propose \textbf{IntraStyler}, the first 3D unpaired image 
translation method for cross-modality MRI domain adaptation that 
synthesizes diverse target domain styles without any predefined 
sub-domain labels.
    
    \item We design a contrastive learning objective to extract 
    per-image style embeddings from 3D MRI that are disentangled 
    from anatomical content, enabling fine-grained style conditioning and smooth style interpolation.
    
    \item Evaluated on the CrossMoDA challenge benchmark, IntraStyler generates more diverse synthetic styles and achieves superior downstream segmentation 
    performance compared to the state-of-the-art method.
\end{itemize}


\begin{figure}[t]
\includegraphics[width=1\columnwidth]{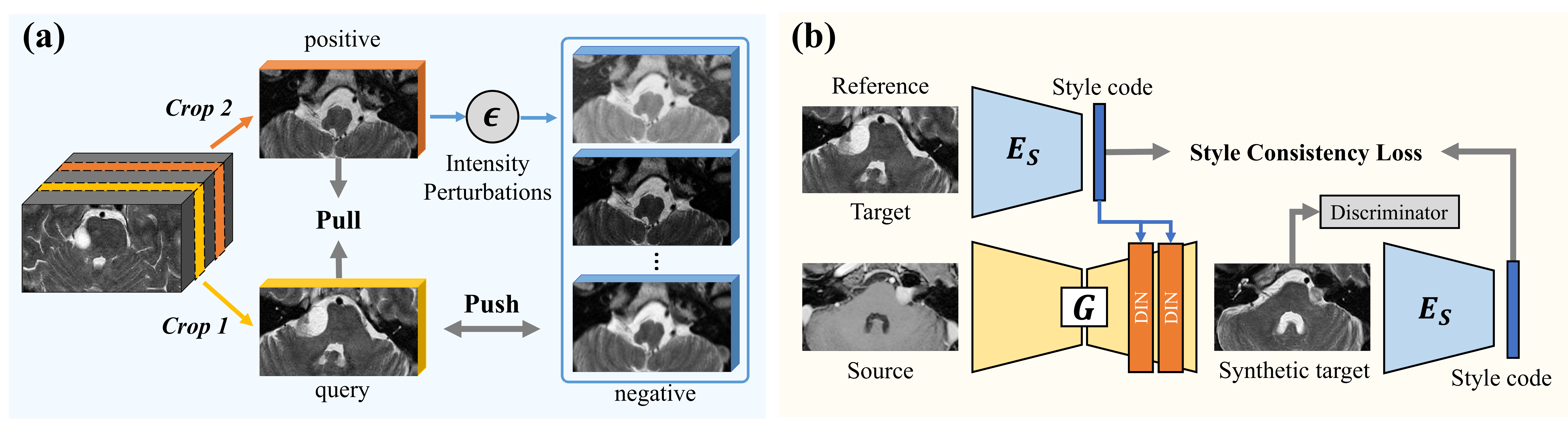}
\centering
\caption{\textbf{(a)} Contrastive learning setup for style encoder $E_S$: given a query, 
the positive shares its style but differs in anatomy, while negatives 
share the positive's anatomy but have different styles.
\textbf{(b)} IntraStyler: the style encoder $E_S$ is jointly trained 
with synthesis backbone $G$, where the style embedding extracted from 
a reference image is used to condition $G$ to synthesize the 
corresponding style.}
\label{fig2}
\end{figure} 

\section{Methods}
As shown in Fig.~\ref{fig2}, IntraStyler consists of two components: (1) a style encoder trained by contrastive learning to extract style-only features from target domain images (Sec.~\ref{contrastive}), and
(2) a synthesis network conditioned on the extracted style embedding for 
controllable image translation (Sec.~\ref{din}).

\subsection{Style Extraction via Contrastive Learning}\label{contrastive}
We aim to train a 3D style encoder $E_S$ that is sensitive to style 
changes but invariant to anatomical content. Our key intuition is that by training on artificially simulated style variations, the encoder learns a general notion of style that transfers to discriminating real MRI appearances without 
requiring any style annotations. As shown in 
Fig.~\ref{fig2}(a), we define \textit{query} as a 3D patch randomly 
cropped from a target domain image, \textit{positive} as another 
patch from the same image at a different location, and 
\textit{negatives} as intensity-perturbed versions of the positive. 
Thus, positive and negatives share the same anatomy but differ in 
style, while only the positive shares the style of the query.

Let $\mathcal{X}$ and $\mathcal{Y}$ denote the source and target 
domains, and let $y$ and $y^+$ denote the query and positive. The 
perturbation function $\epsilon(\cdot)$ is randomly sampled from 
a set of intensity transformations (e.g., contrast adjustment, 
Gaussian smoothing), giving negative $y^- = \epsilon(y^+)$. The 
query, positive, and $N$ negatives are passed through $E_S$ to 
obtain $K$-dimensional style vectors $v$, $v^+\!\in\mathbb{R}^K$ 
and $v^-\!\in\mathbb{R}^{N\times K}$, which are $\ell_2$-normalized 
onto a unit sphere so that similarity reduces to the dot product. 
The contrastive objective is cast as an $(N{+}1)$-way classification 
problem, with cross-entropy loss maximizing the probability of 
selecting the positive over all negatives:

\begin{equation}
 L_{\text{style}}(v,v^{+},v^{-})=-log[\frac{exp(v\cdot v^{+}/\tau)}{exp(v\cdot v^{+}/\tau)+\sum_{n=1}^{N}exp(v\cdot v_{n}^{-}/\tau)}]
\end{equation}

\noindent{where $\tau=0.01$ is the temperature scaling factor for style similarity.


\subsection{Controllable Synthesis with Extracted Styles}\label{din}

To integrate the extracted style with the synthesis model, we inject 
the style embedding as a condition for controllable synthesis, as shown 
in Fig.~\ref{fig2}(b). Following~\cite{liu2023learning}, we adopt 3D 
QS-Attn~\cite{hu2022qs} as the synthesis backbone, selected for its 
attention mechanism and one-sided design that is computationally 
efficient for 3D tasks. Inspired by adaptive normalization 
methods~\cite{dumoulin2017a,karras2019style,jing2020dynamic,huang2017arbitrary}, 
the style embedding is injected via dynamic instance normalization (DIN) 
layers. The input feature $z$ is first channel-wise normalized as in 
standard IN: $z_{\text{norm}}=\frac{z-\mu}{\sigma}$. The style vector 
$v$ is then passed through a $1{\times}1{\times}1$ convolutional mapping 
layer to produce affine parameters $\gamma_{v}$ and $\beta_{v}$, which 
de-normalize the feature as $z_{\text{out}}=\gamma_{v}z_{\text{norm}}+\beta_{v}$. 
In practice, the last two IN layers of the decoder are replaced with 
DIN layers.

To further encourage the consistency between the styles of reference image and the generated image, we use $E_{s}$ to extract the style embedding of the generated image $G(x)$ and introduce a style consistency loss:

\begin{equation}
    L_{\text{con}}= -sim(E_{S}(y), E_{S}(G(x)))
\end{equation}

\noindent where $x\in\mathcal{X}$ is the source domain input image, $G$ is the synthesis network, and $sim(\cdot)$ is the dot product, since the style embeddings are unit vectors.

\subsubsection{Training Objective.} Instead of training style encoder and synthesis network sequentially, we find these two networks can be jointly optimized in an end-to-end manner. The overall training objective of IntraStyler is expressed as: 

\begin{equation}
L_{\text{IntraStyler}} = 
\underbrace{L_{\text{adv}} + \lambda_{\text{NCE}}L_{\text{PatchNCE}}}_{\scriptstyle\text{QS-Attn}\footnotemark}
\;+\;
\underbrace{\lambda_{\text{style}}L_{\text{style}}}_{\scriptstyle\text{style discrimination}}
\;+\;
\underbrace{\lambda_{\text{con}}L_{\text{con}}}_{\scriptstyle\text{style consistency}}
\end{equation}
\footnotetext{$L_{\text{adv}}$ and $L_{\text{PatchNCE}}$ are losses from CUT~\cite{park2020contrastive}, inherited by QS-Attn~\cite{hu2022qs}. See \url{https://github.com/sapphire497/query-selected-attention}.}

\begin{figure}[t]
\includegraphics[width=1\columnwidth]{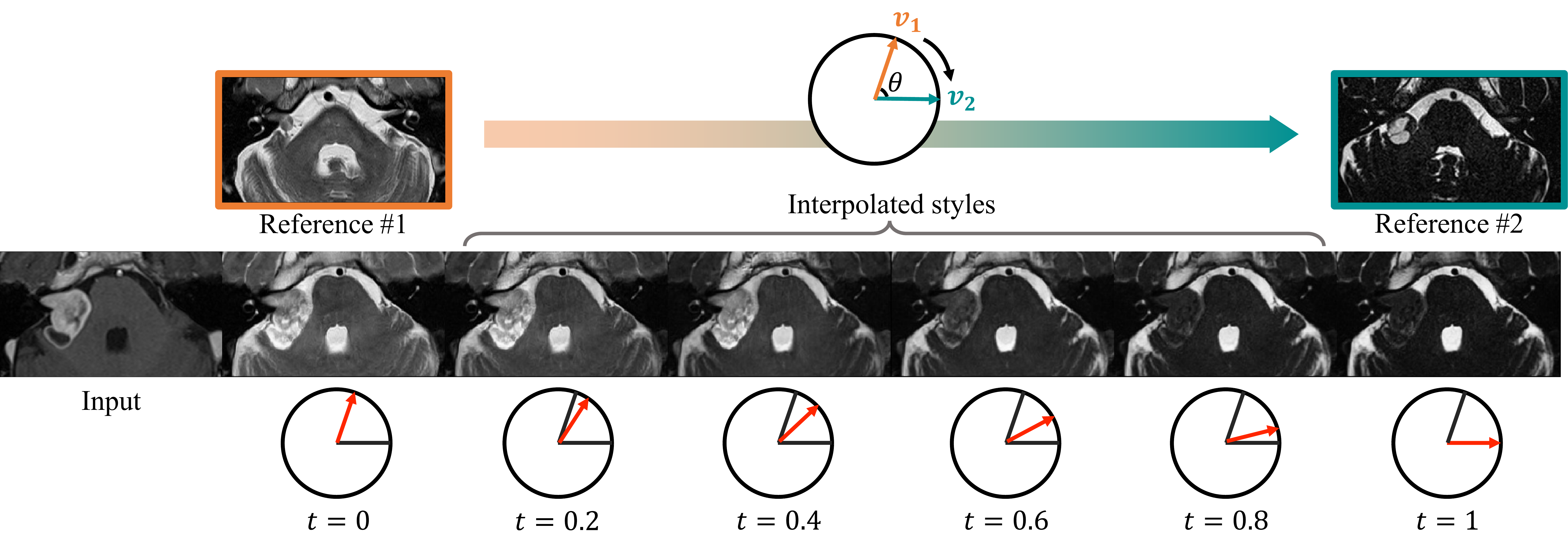}
\centering
\caption{As the interpolation parameter $t$ varies from 0 to 1 (red arrow rotates clockwise), the \textit{appearance} of the synthesized image smoothly transitions between the styles of two reference images, while the anatomical content remains determined by the source input.} \label{fig3}
\end{figure} 

\subsection{Style Interpolation with SLERP}\label{slerp}
Beyond a single reference image, IntraStyler supports style synthesis 
conditioned on two reference images via style interpolation. Standard 
linear interpolation between unit vectors produces intermediate vectors 
of varying magnitude, causing inconsistent style intensities. We therefore 
adopt spherical linear interpolation 
(SLERP)~\cite{jafari2014spherical}, which preserves unit norm throughout 
interpolation and ensures perceptually uniform style transitions. As shown 
in Fig.~\ref{fig3}, the interpolation parameter $t \in [0, 1]$ provides 
explicit control over the degree of style blending between the two 
references. Given unit-norm style embeddings $v_0$ and $v_1$ with angle $\theta = \arccos(v_0 \cdot v_1)$ between them, the interpolated style is computed as:

\begin{equation}
    \text{SLERP}(v_{0}, v_{1}; t)=\frac{\sin((1-t)\theta)}{\sin(\theta)}v_{0}+\frac{\sin(t\theta)}{\sin(\theta)}v_{1}
\end{equation}

\subsection{Implementation Details}

We implement IntraStyler based on the 1st place CrossMoDA solution~\cite{liu2023learning} by 
incorporating our proposed style encoder. 
We set $\lambda_{\text{style}} = \lambda_{\text{con}} = 5$ and the remaining loss weights follow the baseline implementation~\cite{liu2023learning}. 
The style vector dimension is set to $K = 256$. For contrastive 
learning, negative samples are constructed by applying one of the 
following intensity transformations to the positive patch: random 
contrast adjustment, Gaussian smoothing, Gaussian noise, bias field 
corruption, or a random mixture of all four. These transformations simulate the dominant physical sources of MRI appearance variation (e.g., B1 inhomogeneity, sequence parameters, thermal noise), enabling the encoder to learn a broad notion of style that generalizes to real acquisition differences (as shown in Fig.~\ref{fig4}). For each training iteration, $N = 8$ negative samples are generated by independently sampling a random transformation for each.


\section{Experiments and Results}
We evaluate IntraStyler on the CrossMoDA challenge benchmark~\cite{challenge}, which consists of 226 labeled ceT1 MRIs 
(source domain) and 295 unlabeled T2 MRIs (target domain), with 
segmentation masks for cochlea and intra- and extra-meatal components 
of vestibular schwannoma (VS). Data were collected from multiple institutes 
across scanners from four manufacturers (Siemens, Philips, GE, 
Hitachi), field strengths of 1.0T, 1.5T, and 3.0T, and multiple 
acquisition sequences, 
resulting in substantial intra-domain variability. This dataset is organized into only 3 coarse institute-based subsets, where the third subset itself aggregates data from multiple centers. Within each subset, images still vary substantially across scanner types and acquisition sequences, 
making institute labels an unreliable proxy for appearance variation. The test set consists of 96 T2 MRIs drawn from this 
heterogeneous multi-institute, multi-scanner cohort. Next, we investigate 
three key questions to validate the core components of IntraStyler:


\begin{figure}[t]
\includegraphics[width=1\columnwidth]{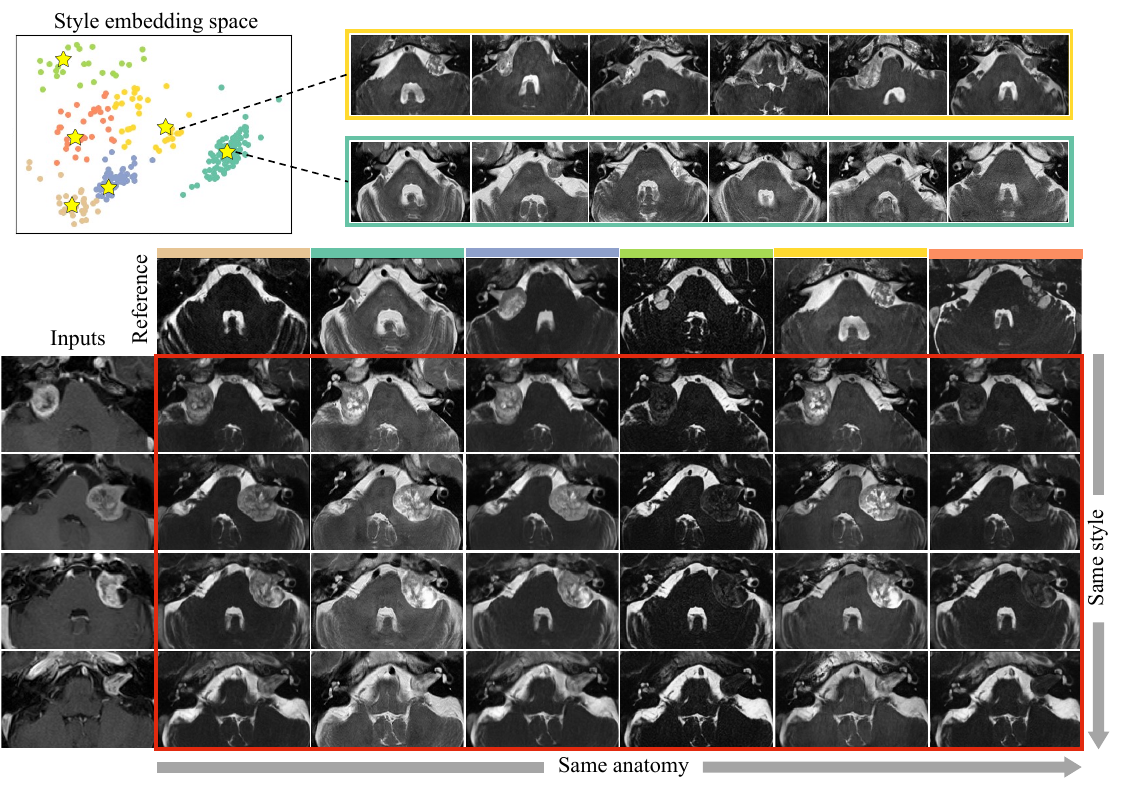}
\centering
\caption{\textbf{Top left}: the style embedding space of all target domain images. Clustering is done by K-means. \textbf{Top right}: visualization of samples from two clusters. Images within the same cluster share consistent styles despite varying anatomies, while images across clusters exhibit clearly distinct appearances. \textbf{Bottom}: source domain images (left column) synthesized to target domain using different reference images (colors of the references correspond to the cluster colors). Synthesis results are shown in the red.} \label{fig4}
\end{figure}

\textbf{Q1. Does contrastive learning extract 
style-only information from real-world heterogeneous MRIs?} To assess our style embeddings, we use the style encoder to project all target domain images into the style embeddings. Then we use K-means to find the clusters of similar style embeddings and visualize them with PCA. The number of clusters is determined by silhouette analysis. As shown in Fig.~\ref{fig4} (top), images within the same cluster share consistent styles despite varying anatomies, while 
images across clusters exhibit clearly distinct appearances, confirming that the learned embeddings capture style independently of anatomy. Notably, although the style encoder is trained only on artificially perturbed styles, it generalizes to 
discriminate real MRI styles.

\textbf{Q2. Do synthesized images follow the style of the reference 
image?} For style synthesis evaluation (Fig.~\ref{fig4}, bottom), four source domain images are translated using six reference images, which are selected as the most representative sample of each cluster by ~\cite{hacohen2022active}. We can observe that (1) column-wise, synthesized images are consistent with their reference style. (2) row-wise, the same source image produces visually distinct outputs across references while preserving anatomy. These results demonstrate that IntraStyler enables fine-grained style control without predefined sub-domains.

\definecolor{intracol}{RGB}{230,240,255}

\begin{table}[t]
\centering

\caption{Quantitative results on segmentation performance.
\textit{Subdomain-free} indicates whether the method requires predefined sub-domain labels.
ASSD follows the challenge evaluation protocol~\cite{challenge}, where a failure is defined 
as the target structure being completely missed, and 350\,mm is assigned to all failure cases.
\#Fail reports the number of failures per method.
Shaded cells indicate a statistically significant difference from IntraStyler 
(Wilcoxon signed-rank, $p{<}0.05$). \textbf{Bold} = best per row.}

\label{tab:results}
\setlength{\tabcolsep}{4pt}
\renewcommand{\arraystretch}{0.95}
\footnotesize
\begin{tabular}{llcccc}
\toprule
 & & NoVar & MultiNets & Dynamic~\cite{liu2023learning} & IntraStyler \\
\midrule
\multicolumn{2}{l}{Diverse styles} & \Xmark & \Cmark & \Cmark & \Cmark \\
\multicolumn{2}{l}{Unified model} & \Cmark & \Xmark & \Cmark & \Cmark \\
\multicolumn{2}{l}{Subdomain-free} & \Cmark & \Xmark & \Xmark & \Cmark \\
\midrule
\multirow{3}{*}{\rotatebox{90}{Dice$\uparrow$}} & Intra-VS & \cellcolor{intracol}0.08{\scriptsize$\pm$0.14} & 0.63{\scriptsize$\pm$0.24} & \cellcolor{intracol}0.57{\scriptsize$\pm$0.26} & \textbf{0.69{\scriptsize$\pm$0.12}} \\
 & Extra-VS & \cellcolor{intracol}0.10{\scriptsize$\pm$0.22} & \cellcolor{intracol}0.70{\scriptsize$\pm$0.29} & \cellcolor{intracol}0.61{\scriptsize$\pm$0.34} & \textbf{0.81{\scriptsize$\pm$0.12}} \\
 & Cochlea & \cellcolor{intracol}0.81{\scriptsize$\pm$0.03} & \cellcolor{intracol}0.82{\scriptsize$\pm$0.07} & \cellcolor{intracol}0.82{\scriptsize$\pm$0.04} & \textbf{0.83{\scriptsize$\pm$0.03}} \\
\midrule
\multirow{4}{*}{\rotatebox{90}{ASSD$\downarrow$}} & Intra-VS & \cellcolor{intracol}219.38{\scriptsize$\pm$169.51} & 22.50{\scriptsize$\pm$85.01} & \cellcolor{intracol}30.46{\scriptsize$\pm$96.89} & \textbf{4.35{\scriptsize$\pm$35.67}} \\
 & Extra-VS & \cellcolor{intracol}222.41{\scriptsize$\pm$168.24} & 28.92{\scriptsize$\pm$95.53} & \cellcolor{intracol}53.11{\scriptsize$\pm$125.16} & \textbf{0.60{\scriptsize$\pm$0.31}} \\
 & Cochlea & \cellcolor{intracol}0.24{\scriptsize$\pm$0.14} & 0.40{\scriptsize$\pm$1.75} & \cellcolor{intracol}0.23{\scriptsize$\pm$0.14} & \textbf{0.21{\scriptsize$\pm$0.12}} \\
 & Boundary & \cellcolor{intracol}269.88{\scriptsize$\pm$147.50} & 36.88{\scriptsize$\pm$106.98} & \cellcolor{intracol}64.96{\scriptsize$\pm$136.10} & \textbf{0.65{\scriptsize$\pm$0.33}} \\
\midrule
\multirow{3}{*}{\rotatebox{90}{\#Fail$\downarrow$}} & Intra-VS & 60 & 6 & 8 & \textbf{1} \\
 & Extra-VS & 55 & 7 & 13 & \textbf{0} \\
 & Cochlea & 0 & 0 & 0 & 0 \\
\bottomrule
\end{tabular}
\end{table}


\textbf{Q3. Does style diversity benefit downstream 
segmentation?} We investigate the impact of different synthesis methods on the downstream segmentation task. For fair comparison, we compare four synthesis strategies that differ only in how 
intra-domain variability is handled, while keeping the downstream segmentation model fixed (nnU-Net~\cite{isensee2021nnu}). We are aware of the existence of other DA approaches that do not require explicit image synthesis, e.g., feature-level alignment. \textit{Comparison against these methods is beyond the scope of our study} because (1) our major contribution lies specifically in style diversification during image translation and (2) IntraStyler is implemented based on the challenge-winning method~\cite{liu2023learning}, which already 
outperformed other DA strategies.

We compare the following synthesis strategies: 
(1) \textbf{Dynamic}~\cite{liu2023learning}: the 1st place solution of the 
CrossMoDA challenge, which generates diverse styles by conditioning 
a unified network on 3 predefined institute-based 
sub-domains, represented as one-hot codes (e.g., $[1,0,0]$).
(2) \textbf{MultiNets}: shares the same backbone as Dynamic but 
replaces the unified design with separate synthesis networks trained 
for each of the 3 institute-based sub-domains, and thus learns an 
easier task per network at the cost of reduced efficiency.
(3) \textbf{NoVar}: uses the same backbone as Dynamic but without 
any intra-domain modeling, serving as a lower bound to quantify 
the impact of ignoring target domain variability. 
(4) \textbf{IntraStyler}: to avoid selecting reference images with 
repetitive styles, we project all target domain images into the 
learned style embedding space and apply K-means clustering with 
silhouette analysis to identify meaningful style groups. Within 
each cluster, the most representative image is selected using the 
diversity-based method of~\cite{hacohen2022active}. This yields 7 reference images that capture the majority of style 
diversity in the target domain. Compared to sub-domain-based methods 
that produce only 3 distinct styles, IntraStyler generates a richer 
set of synthetic target domain images for downstream segmentation training.

We evaluate segmentation results using Dice and Average Symmetric 
Surface Distance (ASSD) across Intra-VS, Extra-VS, and Cochlea, with 
Boundary ASSD (i.e., the distance between the intra- and extra-VS 
boundary) additionally reported. Following the challenge evaluation 
protocol~\cite{challenge}, a failure is defined as the target structure 
being completely missed, and a 350\,mm penalty is assigned to all 
failure cases in the ASSD computation. We report the quantitative results in Tab.~\ref{tab:results} with statistical significance 
between each baseline and IntraStyler using the Wilcoxon signed-rank test with Bonferroni correction ($p{<}0.05$).


NoVar performs significantly worse than IntraStyler across all 
structures, with 60 and 55 complete failures on Intra-VS and 
Extra-VS respectively, confirming that data augmentation alone is insufficient 
to address real-world intra-domain variability and model-based style 
synthesis is necessary. 
Notably, all methods achieve zero Cochlea failures, suggesting 
that intra-domain variability primarily affects VS segmentation. 
Among sub-domain-based methods, MultiNets is the strongest 
competitor yet still produces substantially more failures 
(6 vs.\ 1 on Intra-VS, 7 vs.\ 0 on Extra-VS) with significantly 
lower Dice on Extra-VS and Cochlea. Dynamic, the prior challenge winning method, similarly underperforms IntraStyler significantly 
across nearly all Dice and ASSD metrics. Both methods share a 
fundamental limitation: institute-level sub-domains are too coarse 
to capture true intra-domain variation, as images from the same 
institute can differ substantially across scanners and protocols. 
IntraStyler overcomes this by automatically discovering fine-grained 
style clusters within the target domain, achieving the best 
performance across all reported metrics.

\section{Discussion and Conclusion}
IntraStyler demonstrates that predefined sub-domain labels are not 
necessary for diverse style synthesis in cross-modality DA. 
By treating each target domain image as an independent style reference, 
the method naturally scales to any degree of intra-domain heterogeneity 
without additional prior information. IntraStyler has two limitations. \underline{First}, the intensity perturbation functions for negative sample construction are designed for MR and and may require modality-specific tuning for other imaging modalities, such as CT and PET. \underline{Second}, the global style vector from average pooling discards spatial information, 
preventing control of local style variations such as tumor-specific 
textures. Our global style conditioning could be combined with local tumor augmentation approaches~\cite{salle2024cross} to further improve robustness. This work opens several future research directions. \underline{First}, the 
learned style embedding space exhibits meaningful clustering 
without any supervision, suggesting its 
potential utility for unsupervised scanner harmonization and 
image retrieval. \underline{Second}, while IntraStyler is currently built 
on a GAN-based backbone, the style conditioning mechanism is 
architecture-agnostic and its integration with diffusion models 
or flow matching models is a promising direction for 
higher-fidelity synthesis. We believe IntraStyler provides a practical foundation for intra-domain variability in real-world DA settings where sub-domain annotations are often unavailable.

\section*{Acknowledgements}
This work was supported in part by the National Institutes of Health grants R01HD109739 and T32EB021937, as well as National Science Foundation grant 2220401. This work was also supported by the Advanced Computing Center for Research and Education (ACCRE) of Vanderbilt University.

\noindent\textbf{Disclaimer.} For research purposes only. Not for clinical use. This prototype is still under development and not yet commercially available. Future commercial availability cannot be guaranteed.

\bibliographystyle{splncs04}
\bibliography{references.bib}
\end{document}